\documentclass[conference]{IEEEtran}
\usepackage{cite}

\ifCLASSINFOpdf
   \usepackage[pdftex]{graphicx}
   \graphicspath{{../pdf/}{../jpeg/}}
   \DeclareGraphicsExtensions{.pdf,.jpeg,.png}
\else
   \usepackage[dvips]{graphicx}
   \DeclareGraphicsExtensions{.eps}
\fi
\usepackage[cmex10]{amsmath}
\usepackage{algorithmic}
\algsetup{linenodelimiter=.}

\usepackage[caption=false,font=footnotesize]{subfig}

\usepackage{amssymb}

\usepackage{multirow}

\begin{document}
%
\title{BEBP: An Poisoning Method Against Machine Learning Based IDSs}

\author{\IEEEauthorblockN{Pan Li, Qiang Liu, Wentao Zhao, Dongxu Wang, Siqi Wang}
\IEEEauthorblockA{College of Computer\\National University of Defense Technology\\
Changsha, Hunan, China 410073\\
Email: \{lipan16, qiangliu06, wtzhao, wangdongxu15\}@nudt.edu.cn, wangsiqi10c@gmail.com}}


%


\maketitle

\begin{abstract}
In big data era, machine learning is one of fundamental techniques in intrusion detection systems (IDSs).
Poisoning attack, which is one of the most recognized security threats towards machine learning-based IDSs, injects some adversarial samples into the training phase, inducing data drifting of training data and a significant performance decrease of target IDSs over testing data. In this paper, we adopt the Edge Pattern Detection (EPD) algorithm to design a novel poisoning method that attack against several machine learning algorithms used in IDSs. Specifically, we propose a boundary pattern detection algorithm to efficiently generate the points that are near to abnormal data but considered to be normal ones by current classifiers. Then, we introduce a Batch-EPD Boundary Pattern (BEBP) detection algorithm to overcome the limitation of the number of edge pattern points generated by EPD and to obtain more useful adversarial samples. Based on BEBP, we further present a moderate but effective poisoning method called chronic poisoning attack. Extensive experiments on synthetic and three real network data sets demonstrate the performance of the proposed poisoning method against several well-known machine learning algorithms and a practical intrusion detection method named FMIFS-LSSVM-IDS.
\end{abstract}


%
\IEEEpeerreviewmaketitle

\section{Introduction}
Currently, intelligent intrusion detection systems (IDSs) generally adopt various machine learning techniques to make decisions regarding the presence of security threats using high performance classifiers, which are selected via learning models and algorithms like support vector machine (SVM), Native Bayes (NB), logistic regression (LR), decision tree and artificial neural networks \cite{ambusaidi2016building}\cite{Kishimoto2011Improving}. For example, the authors in \cite{ambusaidi2016building} proposed an efficient intrusion detection method by combining flexible mutual information based feature selection (FMIFS) and least-square SVM (LSSVM), achieving the state-of-the-art classification performance on widely recognized KDDCUP99, NSL-KDD and Kyoto 2006+ data sets.

Although machine learning has been extensively used for intelligent decision in IDSs, previous works have demonstrated that the technology itself suffers from diverse security threats, e.g., attacking against spam filtering \cite{Nelson2009Misleading}, malware detection \cite{Biggio2014Poisoning}\cite{Hu:2017wh} and anomaly detection systems \cite{Kloft2011Online}\cite{Rubinstein2009ANTIDOTE}. Basically, security threats towards machine learning can be classified into two categories, i.e., exploratory and causative attacks \cite{Barreno2006Can}. Specifically, the exploratory attack exploits the security vulnerabilities of learning models to deceive the resulting classifiers without affecting their training phase. For example, adversaries generate some customized adversarial samples\footnote{The terms sample and data point are used interchangeably in this paper for convenience.} to evade the detection of spam filtering \cite{Nelson2009Misleading} and malware detection systems \cite{Hu:2017wh}\cite{Xu2016Automatically}. Considering the great influences of deep neural networks (DNNs) in several application scenarios, e.g., speech recognition, image recognition, natural language processing and autonomous driving, some researchers paid more attention to exploratory attacks against prevailing DNNs \cite{Papernot2017Practical}\cite{Moosavidezfooli2016DeepFool}. On the other hand, the causative attack (also termed as the poisoning attack) shall change training data sets via injecting adversarial samples, inducing influences on the training phase of learning models \cite{Barreno2006Can}. Typically, such adversarial samples are designated by adversaries to have similar features with malicious samples but wrong labels, inducing the change of the training data distribution. Therefore, adversaries can reduce the performance of classification or regression models in terms of accuracy. Since the training data in practical machine learning based systems are protected with high confidentiality, it is uneasy for adversaries to alter the data themselves. Alternatively, the adversaries are able to exploit the vulnerability that stems from retraining existing machine learning models. Since machine learning based systems in practical usage, e.g., anomaly detection systems \cite{Kloft2011Online}\cite{Rubinstein2009ANTIDOTE}, are generally required to periodically update their decision models to adapt to varying application contexts, the poisoning attack is emerging as a main security threat towards these real systems. Hence, we focus on the latter type of security threats towards machine learning in this paper.

Existing work regarding poisoning attacks mainly falls into poisoning SVMs \cite{Biggio2012Poisoning}, principal component analysis (PCA) \cite{Rubinstein2009ANTIDOTE} via direct gradient methods. However, these attacking methods are not effective to poison other learning models. Recently, a poisoning attack against DNNs was proposed by adopting the concept of generative adversarial network (GAN) \cite{Yang:2017ve}. Label contamination attack (LCA) is another type of poisoning attack against black-box learning models \cite{Zhao:2017jj}. However, LCA made a strong assumption that the adversary had the ability of changing the labels of training data, which was difficult in reality. In addition, some researchers proposed another attaching strategy called model inversion by using the information of system application program interfaces (APIs) \cite{Rosenberg:2017vd}\cite{Tramer:2016um}.

In this paper, we propose a novel poisoning method using the Edge Pattern Detection (EPD) algorithm described in \cite{Li2011Selecting}\cite{Wang2017MST}. Specifically, we propose a boundary pattern detection algorithm to efficiently generate the poisoning data points that are near to abnormal data but regarded as normal ones by current classifiers. After that, we present a Batch-EPD Boundary Pattern (BEBP) detection algorithm to address the drawback of the limited number of edge pattern points generated by conventional EPD and to obtain more useful boundary pattern points. After that, we present a moderate but effective poisoning method based on BEBP, called chronic poisoning attack. Compared to previous poisoning methods, a notable advantage of the proposed poisoning method is that it can poison different learning models such as SVMs (with linear, RBF, sigmoid and polynomial kernels), NB and LR. Extensive experiments on synthetic and three real network data sets, i.e., KDDCUP99, NSL-KDD and Kyoto 2006+, demonstrate the effectiveness of the proposed poisoning method against the above learning models and a practical intrusion detection method named FMIFS-LSSVM-IDS (see \cite{ambusaidi2016building}).

The rest of this paper is organized as follows: Section II presents an adversary model and some assumptions. Then, Section III gives the details of the proposed poisoning method. After that, Section IV evaluates the performance of the proposed method via extensive experiments on synthetic and real network data sets. Finally, Section V concludes this paper.

\section{Adversary Model and Assumptions}
In this section, we present an adversary model and make some proper assumptions from four aspects: goal, knowledge, capability and strategy.

(a) \textbf{The adversarial goal}. Generally speaking, the adversarial goal means the intention of launching attacks, e.g., breaking integrity, availability and user privacy \cite{Barreno2006Can}\cite{Biggio2017Security}. In poisoning attack, integrity violation and availability violation are two dominating goals of an adversary. To be more detailed, the adversary hopes to attack against a learning model and its application performance by poisoning the training phase. Hence, we assume that the adversarial goal is to reduce the accuracy and the detection rate of IDSs.

(b) \textbf{The adversarial knowledge}. To achieve the above goal, an adversary should have some information related to target IDS systems. Thus, the adversarial knowledge is the priori information that the adversary can utilize to design attacking strategies, including learning algorithms, training and testing data sets, extracted features, etc \cite{Barreno2006Can}. Conventional poisoning methods \cite{Kloft2011Online}\cite{Rubinstein2009ANTIDOTE} require full knowledge of target systems, which is not rational in practical usage. Therefore, we make an assumption of limited knowledge that the adversary only know the details of training data.

(c) \textbf{The adversarial capability}. The adversarial capability of launching poisoning attacks contains two points. One is whether or not the adversary can change the labels or the features of training data. The other is how many adversarial samples that the adversary can inject into training data. Accordingly, we make two more assumptions regarding the adversarial capability as follows: the adversary can not change the labels nor modify the features of training data and is able to inject adversarial samples at each time of updating target systems (or retraining learning models).

(d) \textbf{The adversarial strategy}. Based on the assumptions made before, we define the adversarial strategy as
\begin{equation}
\label{eqn1} \text{minimize}\quad P(m|\theta)
\end{equation}
\begin{equation}
\label{eqn2} \text{s.t.}\quad N(\mathcal{D}_a) < \eta N(\mathcal{D}_{tr}),
\end{equation}
where $N(\mathcal{D}_{tr})$ denotes the total number of training samples, $\eta$ is a constant parameter representing the poisoning degree of adversarial samples, $m$ refers to the target learning model that the adversary aims to compromise, $\theta$ means the adversarial knowledge including training data and the output labels after feeding inputs, and $N(\mathcal{D}_a)$ represents the number of adversarial samples that the adversary can inject. Thus, the adversarial goal $P$ is to minimize the performance of the target learning model $m$ under limited knowledge $\theta$ and capability $N(\mathcal{D}_a)$.

\section{Details of the Proposed Batch Poisoning Method}
\subsection{Formulation of Adversarial Sample Generation}
According to the adversary model, an adversary has no knowledge about learning models in machine learning-based IDSs. Hence, the proposed poisoning method can be regarded as a kind of black-box attacks. Since the information about learning models is unknown, the adversary alternatively prefers to inject adversarial samples such that target models can not well fit for the real distribution of training data. Such process is termed as \textit{data drifting} in this paper.

To maximize the effects of data drifting in training data, the best strategy is to generate adversarial samples that are close to the discriminant plane defined by a pretrained decision function $f(\mathbf{x})$. Hence, the black-box poisoning problem can be formally defined by generating a set of adversarial samples $\mathcal{D}_a$ satisfying
\begin{equation}
\label{eqn3} \mathcal{D}_a = \{\mathbf{x}_a | d(\mathbf{x}_a, \mathbf{x}_b) < \varepsilon, f(\mathbf{x}_b)=0 \},
\end{equation}
where $d(\mathbf{x}_1,\mathbf{x}_2)$ is the Euclidean distance between two vectors, and $\varepsilon$ denotes the chosen threshold between an adversarial sample $\mathbf{x}_a\in \mathcal{D}_a$ and the discriminant plane.

\subsection{Boundary Pattern Detection}
As per the formulation of adversarial sample generation, we define \textit{boundary pattern} as the data points that are near to abnormal data but considered as normal ones by classifiers. Thus, the goal of the proposed poisoning method is to generate the boundary pattern, which is then used to shift discriminant plane towards the central of abnormal data during model retraining. Accordingly, we propose a boundary pattern detection (BPD) algorithm using the edge pattern detection (EPD) algorithm \cite{Li2011Selecting}\cite{Wang2017MST} to effectively generate the boundary pattern samples. There are two main steps in BPD as follows:

(a) Detecting the edge pattern points of normal data that are regarded as normal behaviors by IDSs. Given $\mathcal{D}_{nd}$, it is easy to find out the edge pattern points $\mathcal{D}_{ep}$ ($\mathcal{D}_{ep}\subset \mathcal{D}_{nd}$) by applying the EPD algorithm \cite{Li2011Selecting}. Moreover, we calculate the normal vector with respect to each edge point to obtain the direction of departing from $\mathcal{D}_{nd}$ with the fastest speed \cite{Wang2017MST}. Let $\mathcal{N}$ denote the set of all normal vectors with respect to $\mathcal{D}_{ep}$.

(b) Generating the boundary pattern by shifting the edge pattern points outwards. Although these edge pattern points locate at the exterior surface of $\mathcal{D}_{nd}$, they may be far from the discriminant plane $f(\mathbf{x})$. Hence, we perform the following two operations based on $\mathcal{D}_{ep}$ and $\mathcal{N}$: Firstly, selecting an edge pattern point $\mathbf{x}_{ep}\in \mathcal{D}_{ep}$ and corresponding normal vector $\mathbf{n}_{ep}\in \mathcal{N}$. Then, shifting $\mathbf{x}_{ep}$ outwards along the direction of $\mathbf{n}_{ep}$ until the generated data points are near to the discriminant plane of classifiers. The data shifting is formally defined by
\begin{equation}
\label{eqn4} \mathbf{x}_a^{i} = \mathbf{x}_a^{i-1} + k_{i-1}\cdot \lambda_{i-1}\cdot \mathbf{n}_{ep},
\end{equation}
where
\begin{equation}
\label{eqn5} \begin{cases}
k_i=1, \lambda_i = \lambda_{i-1}, & \text{if}\ f(\mathbf{x}_a^{i-1})\ \text{is normal} \\
k_i = -1, \lambda_i = \lambda_{i-1}/3, & \text{otherwise}
\end{cases}
\end{equation}

\begin{figure}[b]
\centering
\begin{algorithmic}[1]
    \STATE \textbf{Input}: An edge pattern point $\mathbf{x}_{ep}$ and corresponding normal vector $\mathbf{n}_{ep}$, target learning model $M$, $m$, $\lambda$
    \STATE \textbf{Output}: A boundary pattern $\mathcal{D}_{bp}$ generated from $\mathbf{x}_{ep}$
    \STATE Initialize $\lambda_0 = \lambda$, $\mathbf{x}_a^0 = \mathbf{x}_{ep}$, $\mathcal{D}_{bp} = \emptyset$;
    \FOR{$i=0,\cdots,m-1$}
        \IF{$f_M(\mathbf{x}_{a}^i) == N$}
            \IF{($\exists \mathbf{x}_b, f_M(\mathbf{x}_b)=0$ \AND $d(\mathbf{x}_a^i,\mathbf{x}_b) < \varepsilon$)}
                \STATE $\mathcal{D}_{bp} = \mathcal{D}_{bp} \bigcup \{\mathbf{x}_{a}^i\}$;
            \ENDIF
            \STATE $\mathbf{x}_{a}^{i+1} = \mathbf{x}_{a}^i + \lambda_i\cdot \mathbf{n}_{ep}$;  $\lambda_{i+1}=\lambda_i$;
        \ELSE
            \STATE $\mathbf{x}_{a}^{i+1} = \mathbf{x}_{a}^i - \lambda_i\cdot \mathbf{n}_{ep}$;  $\lambda_{i+1} = \lambda_i/3$;
        \ENDIF
    \ENDFOR
\end{algorithmic}
\caption{Pseudo code of the boundary pattern detection algorithm}\label{alg1}
\end{figure}

The pseudo code of the BPD algorithm is shown in Fig. \ref{alg1}, where $m$ is the maximal number of iterations, $\lambda$ means the initial shifting step size, $\mathbf{x}_{ep}$ and $\mathbf{n}_{ep}$ represent the selected edge pattern point and corresponding normal vector, respectively. In particular, we first shift $\mathbf{x}_{ep}$ outwards along the direction of its normal vector $\mathbf{n}_{ep}$ according to equations (\ref{eqn4}) and (\ref{eqn5}), where $\mathbf{x}_a^i$ and $\lambda_i$ determine the generated adversarial sample and the shifting step size in the $i$th iteration. Note that $\mathbf{x}_a^0 = \mathbf{x}_{ep}$. Furthermore, the output of a target learning model $M$ ($f_M(\mathbf{x})$) with respect to an input sample $\mathbf{x}$ falls into $\{N, A\}$ representing \textit{Normal} and \textit{Abnormal}, respectively. Finally, we select valid adversarial samples (i.e. boundary pattern points) according to the equation (\ref{eqn3}). For simplicity, $\varepsilon$ is set to $\lambda$.

\subsection{Batch-EPD Boundary Pattern Detection}
\begin{figure}[t]
\centering
\begin{algorithmic}[1]
    \STATE \textbf{Input}: A training data set $\mathcal{D}_{tr}$, target learning model $M$, maximal number of iterations $m$, shifting step size $\lambda$, batch size $L$;
    \STATE \textbf{Output}: Generated adversarial samples $\mathcal{D}_a$;
    \STATE Select the training data $\mathcal{D}_{tr}^{(N)}$ with normal labels from $\mathcal{D}_{tr}$;
    \STATE Initialize $\mathcal{D}_a = \emptyset$, $k=size(\mathcal{D}_{tr}^{(N)})/L$;
    \FOR{$i=1,\cdots,k$}
        \STATE Randomly select $L$ samples from $\mathcal{D}_{tr}^{(N)}$, which is denoted by $\mathcal{D}_i^{(N)}$;
        \STATE Calculate $\mathcal{D}_{ep}$ and corresponding $\mathcal{N}$ regarding $\mathcal{D}_i^{(N)}$ using EPD;
        \FOR{$\mathbf{x}_{ep}\in \mathcal{D}_{ep}$}
            \STATE Calculate $\mathcal{D}_{bp}$ using BPD with inputs of $\mathbf{x}_{ep}$, $\mathbf{n}_{ep}$, $m$, $M$ and $\lambda$;
            \STATE $\mathcal{D}_a = \mathcal{D}_a \bigcup \mathcal{D}_{bp}$;
        \ENDFOR
    \ENDFOR
\end{algorithmic}
\caption{Pseudo code of the Batch-EPD boundary pattern detection algorithm}\label{alg2}
\end{figure}

Although the aforementioned BPD algorithm can effectively generate the boundary pattern, it is constrained by the limited number of edge pattern points, especially for those data sets with sparse edge points. Hence, we further introduce a Batch-EPD method, which is able to directly obtain more valid adversarial samples near to the discriminant boundary of learning models. The main idea of Batch-EPD is as follows: At the first stage, we randomly select $k$ subsets $\mathcal{D}_1^{(N)}, \mathcal{D}_2^{(N)}, \cdots, \mathcal{D}_k^{(N)}$ from the training data $\mathcal{D}_{tr}^{(N)}$ with \textit{Normal} labels. Then, we utilize the conventional EPD algorithm to calculate edge pattern points and corresponding normal vectors with respect to each subset $\mathcal{D}_i^{(N)}$ ($i=1, 2, \cdots, k$). Note that some edge pattern points generated by Batch-EPD may locate at inner data points of $\mathcal{D}_{tr}^{(N)}$. However, the proposed BPD algorithm can still shift these inner points to the discriminant boundary. Fig. \ref{alg2} shows the pseudo code of the proposed Batch-EPD boundary pattern (BEBP) detection algorithm.

\begin{figure}[t]
\centering
\includegraphics[width=2.8in]{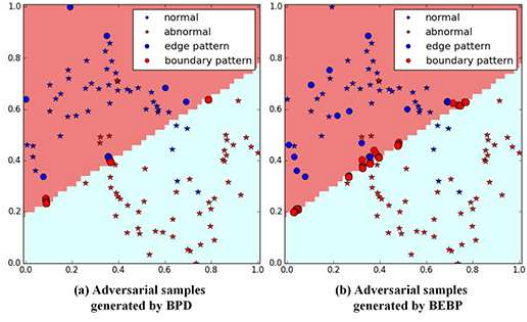}
\caption{Comparative results on a synthetic data set between BPD and BEBP} \label{fig-comp-BPD-BEBP}
\end{figure}

To demonstrate the improvement of BEBP comparing to BPD, Fig. \ref{fig-comp-BPD-BEBP} illustrates comparative results on a synthetic data set, where blue (red) stars are normal (abnormal) samples, blue and red solid circles refer to edge pattern points and generated adversarial samples, respectively.

\subsection{Chronic Poisoning Attack Using BEBP}
Based on the aforementioned BEBP algorithm, we now present a moderate but effective poisoning method against learning models, called \textit{chronic poisoning attack}. Similar to the boil frog poisoning attack proposed in \cite{Rubinstein2009ANTIDOTE}, the proposed chronic poisoning attack using BEBP is also a long-term poisoning method, which changes the distribution of training data in each time of updating learning models. By gradually injecting adversarial samples, which are classified as normal samples and locate near to the discriminant boundary defined by a pretrained model, the boundary of the updated model after retraining over the corrupted training data will move towards the centre of abnormal data points. As a result, the performance of IDSs detecting abnormal samples significantly decreases after several rounds of poisoning. Fig. \ref{alg3} shows the pseudo code of the chronic poisoning attack using BEBP, where $\mathcal{D}_{tr}^{(i)}$ and $M_i$ refer to the training data and the pretrained model at the $i$th round of poisoning, respectively.
\begin{figure}[h]
\centering
\begin{algorithmic}[1]
    \STATE \textbf{Input}: An initial training data set $\mathcal{D}_{tr}^{(0)}$, an initial learning model $M_0$, number of poisoning rounds $r$
    \FOR{$i=0,\cdots,r-1$}
        \STATE Generate adversarial samples $\mathcal{D}_a$ using the BEBP algorithm with inputs $\mathcal{D}_{tr}^{(i)}$ and $M_i$;
        \STATE $\mathcal{D}_{tr}^{(i+1)} = \mathcal{D}_{tr}^{(i)}\bigcup \mathcal{D}_a$;
        \STATE Retrain a new model $M_{i+1}$ based on $\mathcal{D}_{tr}^{(i+1)}$;
    \ENDFOR
\end{algorithmic}
\caption{Pseudo code of the chronic poisoning attack using BEBP}\label{alg3}
\end{figure}

\section{Performance Evaluation and Analysis}
In this section, we evaluate the performance of the proposed algorithms by extensive experiments described as follows: Firstly, we examine the attacking effects of the proposed poisoning method against different learning models on synthetic data sets. Then, we evaluate the performance of the proposed method on three real data sets to further demonstrate its strong capability of reducing the detecting performance of multiple learning models. After that, we select a state-of-the-art IDS system, called FMIFS-LSSVM-IDS \cite{ambusaidi2016building}, as the poisoning target and give comparative results between the proposed method and two other baseline methods.

\subsection{Experimental Setup}
\subsubsection{Data Sets}
To demonstrate the performance of the proposed poisoning method without loss of generality, we adopted the synthetic moon data set that was used in \textit{sklearn}\footnote{http://scikit-learn.org}, where 100 synthetic samples were randomly generated with a noise of 0.2. Regarding the real data sets, we chose three public data sets, i.e., KDDCUP99, NSL-KDD and Kyoto 2006+. KDDCUP99 is a well-known benchmark data set for evaluating the performance of IDSs, which contains five categories of samples (one normal and four abnormal). Moreover, each sample has $41$ features. NSL-KDD is a revised version of KDDCUP99, and it has the same numbers of categories and features. Apart from these two widely used data sets, Kyoto 2006+ proposed in \cite{Song2011Statistical} is another recognized data set for performance evaluation. The data set has been collected from honeypots and regular servers that are deployed at the Kyoto University since 2006. Moreover, Kyoto 2006+ contains three types of samples, i.e., normal, known attack and unknown one, and each sample has $24$ features.

Considering that the goal of poisoning attacks is to reduce the performance of IDSs detecting abnormal behaviors, we treat all samples with abnormal labels in each data set as a whole regardless of their specific types of attacks. Similar to FMIFS-LSSVM-IDS, we preprocess and perform data normalization with respect to all samples such that each feature value is normalized into a range of $[0,1]$. To evaluate the effectiveness of the proposed poisoning method, we will use two types of data for performance evaluation, a.k.a. (a) \textit{evaluating data} that are randomly selected from training data, and (b) \textit{official testing data} from public data sets.

\subsubsection{Performance Metrics}
Regarding an IDS system, accuracy and detecting rate are two primary performance metrics. Hence, we also adopt these two metrics in this paper to evaluate the performance reduction of machine learning-based IDSs under the proposed poisoning attack. The accuracy ($ACC$) and the detecting rate ($DR$) with respect to abnormal samples are defined by
\begin{equation}
\label{eqn6} ACC = \frac{TP+TN}{TP+TN+FN+FP}
\end{equation}
\begin{equation}
\label{eqn7} DR = \frac{TP}{TP+FN},
\end{equation}
where true positive ($TP$) is the number of truly abnormal samples that are classified as abnormal ones by IDSs, true negative ($TN$) means the number of truly normal samples that are treated as normal ones, false positive ($FP$) refers to the number of truly normal samples classified as abnormal ones, and false negative ($FN$) represents the number of truly abnormal samples classified as normal ones.

\subsection{Performance of the Proposed Poisoning Method over Synthetic Data Sets}
\begin{figure*}[t]
\centering
\includegraphics[width=4.8in]{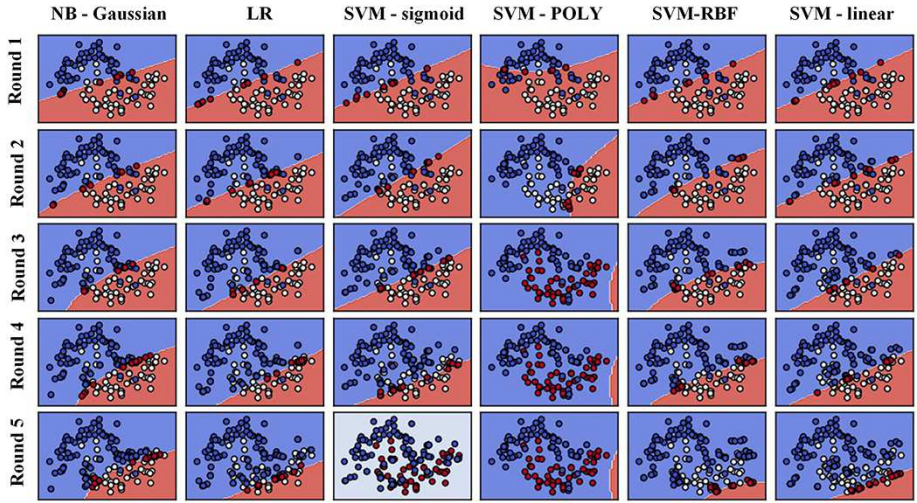}
\caption{Comparative results of five-round poisoning against different learning models on synthetic data sets} \label{fig-comp-five-round-poisoning}
\end{figure*}

\begin{table*}[t]
\renewcommand{\arraystretch}{1.3}
\caption{Summary of Sample Distributions of the Randomly Selected Data Regarding the KDDCUP99 and NSL-KDD Data Sets}
\label{table1} \centering
\begin{tabular}{l l l l l l l}
    \hline
    & \bfseries Data Set & \bfseries NORMAL & \bfseries PROB & \bfseries DOS & \bfseries U2R & \bfseries R2L\\
    \hline
    \multicolumn{1}{l}{\multirow{2}{1.5cm}{KDDCUP99}} & \multicolumn{1}{l}{Training~data} & \multicolumn{1}{l}{2000} & \multicolumn{1}{l}{300} & \multicolumn{1}{l}{3790} & \multicolumn{1}{l}{32} & \multicolumn{1}{l}{350}\\
    \multicolumn{1}{l}{} & \multicolumn{1}{l}{Evaluating~data} & \multicolumn{1}{l}{2000} & \multicolumn{1}{l}{500} & \multicolumn{1}{l}{3900} & \multicolumn{1}{l}{20} & \multicolumn{1}{l}{400}\\
    \multicolumn{1}{l}{\multirow{2}{1.5cm}{NSL-KDD}} & \multicolumn{1}{l}{Training~data} & \multicolumn{1}{l}{2000} & \multicolumn{1}{l}{300} & \multicolumn{1}{l}{3790} & \multicolumn{1}{l}{32} & \multicolumn{1}{l}{350}\\
    \multicolumn{1}{l}{} & \multicolumn{1}{l}{Evaluating~data} & \multicolumn{1}{l}{2000} & \multicolumn{1}{l}{500} & \multicolumn{1}{l}{3900} & \multicolumn{1}{l}{20} & \multicolumn{1}{l}{400}\\
    \hline
\end{tabular}
\end{table*}

To demonstrate the attacking effects of chronic poisoning, we first evaluated the performance of the proposed poisoning method against six different learning models on synthetic data sets. The evaluated models included NB-Gaussian, LR, SVM with a sigmoid kernel (SVM-sigmoid), SVM with a polynomial kernel (SVM-POLY), SVM with a radial basis function kernel (SVM-RBF) and SVM with a linear kernel (SVM-linear). To focus on poisoning itself, we simply used the default values of model parameters as specified in the sklearn tool. Fig. \ref{fig-comp-five-round-poisoning} illustrates the comparative results of five-round poisoning against different learning models, where the blue and white points represent the training data with normal and abnormal labels, respectively. In Fig. \ref{fig-comp-five-round-poisoning}, the read points mean the adversarial samples generated by BEBP, and the discriminant boundary between normal and abnormal samples is shown as the line of separating blue and red regions. Moreover, we would like to highlight that read points in the figures of SVM-sigmoid at the $5$th round and SVM-POLY at the $3$nd-$5$th rounds denote the truly abnormal data. From Fig. \ref{fig-comp-five-round-poisoning}, we can see that no matter what the learning model is, the discriminant boundary gradually moves towards the centre of abnormal data. Accordingly, we clearly figure out that more abnormal points are wrongly classified as normal ones along with an increase of poisoning round.

\subsection{Performance of the Proposed Poisoning Method over real Data Sets}
According to the sample selection method in \cite{Kishimoto2011Improving}, we adopted $6472$ samples as training data and $6820$ samples as evaluating data that were randomly selected from the ``kddcup.data\_10\_percent\_corrected'' (``KDD Train+'') of the KDDCUP99 (NSL-KDD) data set. Table \ref{table1} summarizes the sample distributions of the selected data regarding the KDDCUP99 and NSL-KDD data sets. Similar to \cite{ambusaidi2016building}, we randomly selected $13292$ samples from the traffic data collected during 27-31, August 2009 regarding the Kyoto 2006+ data set.

\begin{figure}[t]
\centering
\includegraphics[width=2.6in]{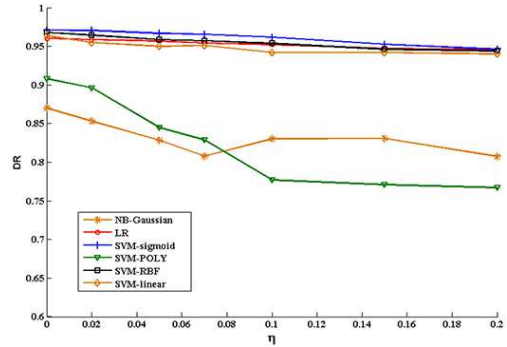}
\caption{Comparative results on NSL-KDD evaluating data with respect to different values of poisoning ratio} \label{fig-comp-diff-ratio}
\end{figure}

As we mentioned before, the parameter $\eta$ controls the poisoning ratio of adversarial samples to normal training data. Hence, it is meaningful to examine the change of poisoning results with different values of $\eta$. For simplicity without loss of generality, we took NSL-KDD as the evaluating data set and carried out a group of experiments with different settings of $\eta$. The comparative results on NSL-KDD evaluating data with respect to different values of poisoning ratio are illustrated in Fig. \ref{fig-comp-diff-ratio}. We can see from Fig. \ref{fig-comp-diff-ratio} that the DR of different learning models tends to decrease with an increase of $\eta$.
\begin{table*}[t]
\renewcommand{\arraystretch}{1.3}
\caption{Comparative Results of $ACC$ on KDDCUP99 under the Proposed Poisoning Attack}
\label{table2} \centering
\begin{tabular}{l l l l l l l}
    \hline
    \bfseries (Evaluating~results;Testing~results) & \bfseries NB & \bfseries LR & \bfseries SVM-sigmoid & \bfseries SVM-POLY & \bfseries SVM-RBF & \bfseries SVM-linear\\
    \hline
    Round~0 & (0.9256;0.8757) & (0.9794;0.9168) & (0.9644;0.9215) & (0.9285;0.919) & (0.981;0.9216) & (0.9809;0.9289)\\
    Round~5 & (0.6102;0.4478) & (0.9311;0.8667) & (0.8825;0.8948) & (0.8517;0.6898) & (0.9091;0.8542) & (0.9304;0.8177)\\
    Round~10 & (0.429;0.3088) & (0.8745;0.8474) & (0.8071;0.6984) & (0.7861;0.6706) & (0.8762;0.8013) & (0.8776;0.7881)\\
    Round~15 & (0.3677;0.247) & (0.8118;0.7089) & (0.7013;0.6398) & (0.3986;0.2989) & (0.7461;0.6303) & (0.7278;0.6679)\\
    \hline
\end{tabular}
\end{table*}
\begin{table*}[t]
\renewcommand{\arraystretch}{1.3}
\caption{Comparative Results of $ACC$ on NSL-KDD under the Proposed Poisoning Attack}
\label{table3} \centering
\begin{tabular}{l l l l l l l}
    \hline
    \bfseries (Evaluating~results;Testing~results) & \bfseries NB & \bfseries LR & \bfseries SVM-sigmoid & \bfseries SVM-POLY & \bfseries SVM-RBF & \bfseries SVM-linear\\
    \hline
    Round~0 & (0.8895;0.7711) & (0.8536;0.7733) & (0.9508;0.781) & (0.892;0.7799) & (0.9576;0.7724) & (0.9578;0.7615)\\
    Round~5 & (0.7726;0.6471) & (0.8822;0.7049) & (0.809;0.6429) & (0.8233;0.6753) & (0.8337;0.6897) & (0.8756;0.6875)\\
    Round~10 & (0.6694;0.5403) & (0.8051;0.646) & (0.7682;0.6034) & (0.7227;0.5162) & (0.7904;0.6552) & (0.7829;0.6222)\\
    Round~15 & (0.6158;0.5164) & (0.7324;0.5563) & (0.5207;0.4683) & (0.3904;0.4445) & (0.6057;0.5155) & (0.6875;0.5125)\\
    \hline
\end{tabular}
\end{table*}
\begin{table*}[t]
\renewcommand{\arraystretch}{1.3}
\caption{Comparative Results of $ACC$ on Kyoto 2006+ under the Proposed Poisoning Attack}
\label{table4} \centering
\begin{tabular}{l l l l l l l}
    \hline
    \bfseries Evaluating~data & \bfseries NB & \bfseries LR & \bfseries SVM-sigmoid & \bfseries SVM-POLY & \bfseries SVM-RBF & \bfseries SVM-linear\\
    \hline
    Round~0 & 0.9541 & 0.9834 & 0.9734 & 0.9315 & 0.9821 & 0.989\\
    Round~5 & 0.6475 & 0.9339 & 0.8984 & 0.869 & 0.9074 & 0.93\\
    Round~10 & 0.6181 & 0.8095 & 0.5457 & 0.4131 & 0.6142 & 0.763\\
    Round~15 & 0.5701 & 0.5422 & 0.4794 & 0.4131 & 0.5362 & 0.5376\\
    \hline
\end{tabular}
\end{table*}

\begin{figure*}[t]
\includegraphics[width=6.8in]{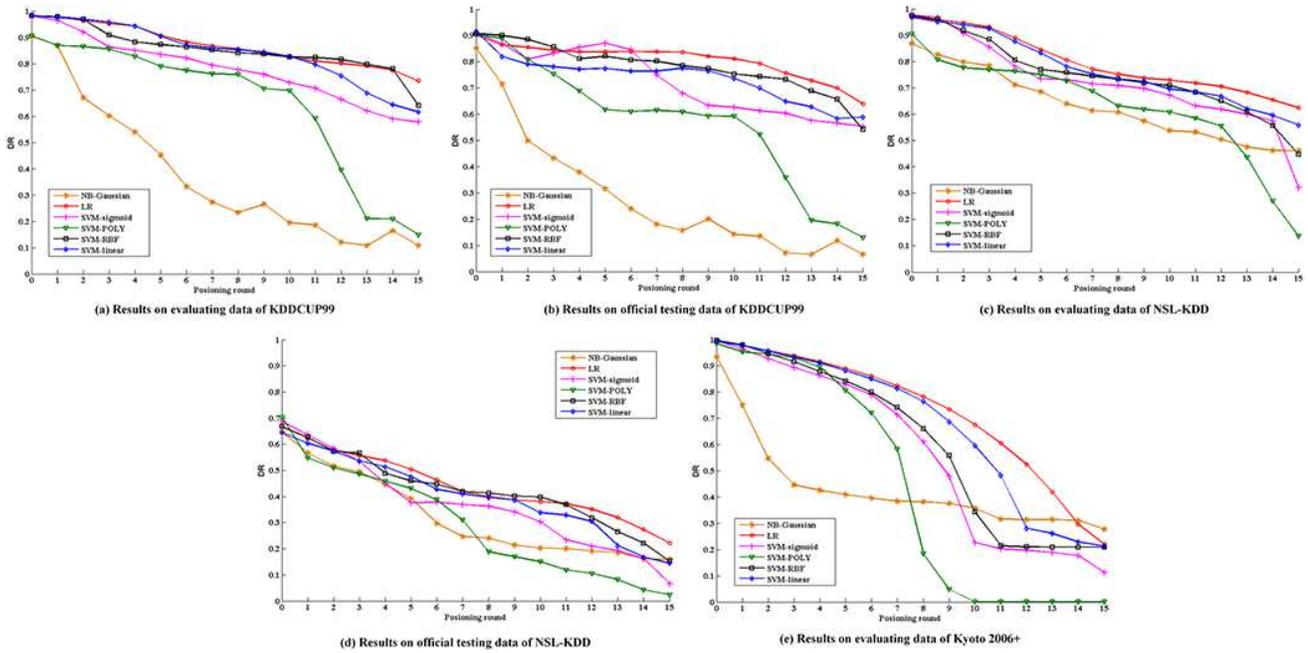}
\caption{Comparative results of $DR$ under the proposed poisoning attack} \label{fig-comp-DR}
\end{figure*}

To further demonstrate the effectiveness of the proposed poisoning method against different learning models, we carried out more experiments on KDDCUP99, NSL-KDD and Kyoto 2006+ data sets. Specifically, we selected the total poisoning round as $15$ in each comparative experiment, and we independently reran poisoning attacks $10$ times to minimize the fluctuation of experimental results brought by random data sampling. Moreover, the poisoning ratio $\eta$ was set to 0.07 in all experiments. The comparative results of $ACC$ and $DR$ under the proposed poisoning attack are given in Tables \ref{table2}--\ref{table4} and Fig. \ref{fig-comp-DR}, respectively. The comparative results on three benchmark data sets demonstrate that both $ACC$ and $DR$ of classifiers detecting abnormal behaviors significantly decrease when the proposed chronic poisoning attack occurs for a long time. Furthermore, the similar changes with respect to different learning models validate that the proposed poisoning method is scalable for attacking black-box detecting models.

\subsection{Comparative Results of Poisoning FMIFS-LSSVM-IDS}
\begin{figure*}[t]
\centering
\includegraphics[width=6in]{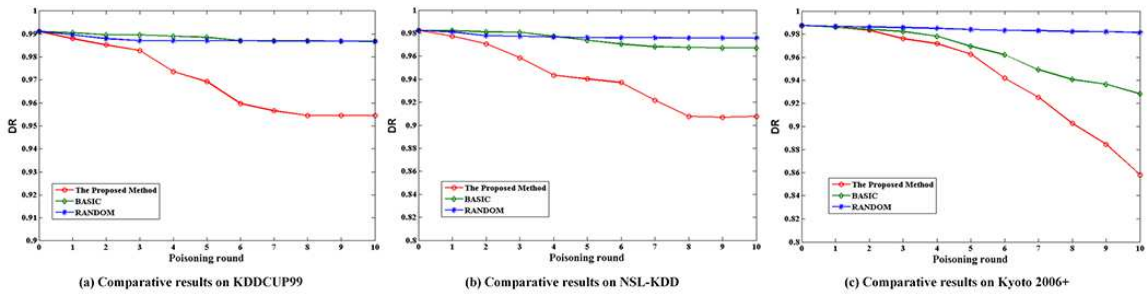}
\caption{Comparative results among different poisoning methods against FMIFS-LSSVM-IDS} \label{fig-comp-FMIFS}
\end{figure*}

In this part, we further demonstrate the performance of the proposed poisoning method against a state-of-the-art IDS based on machine learning named FMIFS-LSSVM-IDS. Here, we select two more poisoning methods as the comparative baselines, i.e., \textit{BASIC} and \textit{RANDOM} \cite{Zhao:2017jj}. In the BASIC method, if $N$ adversarial samples are added into training data, then $N$ normal samples selected from normal training data randomly will also be added. In the RANDOM method, on the other hand, we generate a number of samples with random features. After that, those samples that are classified as the normal ones by FMIFS-LSSVM-IDS are chosen as valid adversarial samples. Finally, some normal samples are randomly selected from normal training data as new added samples as well. Fig. \ref{fig-comp-FMIFS} illustrates the comparative results among different poisoning methods.

We can see from Fig. \ref{fig-comp-FMIFS} that the proposed poisoning method is more effective to reduce the $DR$ of FMIFS-LSSVM-IDS compared with BASIC and RANDOM on all three data sets. These results further demonstrate the advantages of the proposed method to attack against state-of-the-art IDSs.

\section{Conclusion and Future Work}
In this paper, we have proposed a novel poisoning method by using the EPD algorithm. Specifically, we first propose the BPD algorithm to generate adversarial samples that locate near to the discriminant boundary defined by classifiers but are still classified to be normal ones. To address the drawback of limited adversarial samples generated by BPD, we further present the BEBP algorithm to obtain more useful adversarial samples. After that, we introduce a chronic poisoning attack based on BEBP. Extensive experiments on synthetic and real data sets demonstrate the effectiveness of the proposed poisoning method against different learning models and state-of-the-art IDSs, e.g., FMIFS-LSSVM-IDS.

In future, it is worthwhile to do more in-depth studies on the scalability of the proposed poisoning method. Moreover, research on defending against the poisoning method will be an interesting work as well.




\bibliographystyle{IEEEtran}
\bibliography{IEEEabrv,ICC2018}

\end{document}